\documentclass[11pt]{article}

\usepackage{amsmath,amsthm,amsfonts}

\usepackage{amssymb}

\usepackage{epsfig}

\usepackage{rotating}

\usepackage{subfigure}

\usepackage{authblk}

\begin{document}

\title{Enforcing constraints for time series prediction in supervised, unsupervised and reinforcement learning}

\author[]{Panos Stinis}

\affil[]{Advanced Computing, Mathematics and Data Division, Pacific Northwest National Laboratory, Richland WA 99354}

\renewcommand\Affilfont{\itshape\small}

\date {}

\maketitle

\begin{abstract}
We assume that we are given a time series of data from a dynamical system and our task is to learn the flow map of the dynamical system. We present a collection of results on how to enforce constraints coming from the dynamical system in order to accelerate the training of deep neural networks to represent the flow map of the system as well as increase their predictive ability. In particular, we provide ways to enforce constraints during training for all three major modes of learning, namely supervised, unsupervised and reinforcement learning. In general, the dynamic constraints need to include terms which are analogous to memory terms in model reduction formalisms. Such memory terms act as a restoring force which corrects the errors committed by the learned flow map during prediction. 

For supervised learning, the constraints are added to the objective function. For the case of unsupervised learning, in particular generative adversarial networks, the constraints are introduced by augmenting the input of the discriminator. Finally, for the case of reinforcement learning and in particular actor-critic methods, the constraints are added to the reward function. In addition, for the reinforcement learning case, we present a novel approach based on homotopy of the action-value function in order to stabilize and accelerate training. We use numerical results for the Lorenz system to illustrate the various constructions.
\end{abstract}

\section*{Introduction}

Scientific machine learning, which combines the strengths of scientific computing with those of machine learning, is becoming a rather active area of research. Several related priority research directions were stated in the recently published report \cite{doe_sml_report}. In particular, two priority research directions are: (i) how to leverage scientific domain knowledge in machine learning (e.g. physical principles, symmetries, constraints); and (ii) how can machine learning enhance scientific computing (e.g reduced-order or sub-grid physics models, parameter optimization in multiscale simulations). 

Our aim in the current work is to present a collection of results that contribute to both of the aforementioned priority research directions. On the one hand, we provide ways to enforce constraints coming from a dynamical system during the training of a neural network to represent the flow map of the system. Thus, prior domain knowledge is incorporated in the neural network training. On the other hand, as we will show, the accurate representation of the dynamical system flow map through a neural network is equivalent to constructing a temporal integrator for the dynamical system modified to account for unresolved temporal scales. Thus, machine learning can enhance scientific computing. 

We assume that we are given data in the form of a time series of the states of a dynamical system (a training trajectory). Our task is to train a neural network to learn the flow map of the dynamical system. This means to optimize the parameters of the neural network so that when it is presented with the state of the system at one instant, it will predict accurately the state of the system at another instant which is a fixed time interval apart. If we want to use the data alone to train a neural network to represent the flow map, then it is easy to construct simple examples where the trained flow map has rather poor predictive ability \cite{stinis2018}. The reason is that the given data train the flow map to learn how to respond accurately as long as the state of the system is on the trajectory. However, at every timestep, when we invoke the flow map to predict the estimate of the state at the next timestep, we commit an error. After some steps, the predicted trajectory veers into parts of phase space where the neural network has not trained. When this happens, the neural network's predictive ability degrades rapidly. 

One way to aid the neural network in its training task is to provide data that account for this inevitable error. In \cite{stinis2018}, we advanced the idea of using a noisy version of the training data i.e. a noisy version of the training trajectory. In particular, we attach a noise cloud around each point on the training trajectory. During training, the neural network learns how to take as input points from the noise cloud, and map them back to the {\it noiseless} trajectory at the next time instant. This is an {\it implicit} way of encoding a restoring force in the parameters of the neural network (see Section \ref{implicit_error_correction} for more details).  We have found that this modification can improve the predictive ability of the trained neural network but up to a point (see Section \ref{numerical} for numerical results). 

We want to aid the neural network further by enforcing constraints that we know the state of the system satisfies. In particular, we assume that we have knowledge of the differential equations that govern the evolution of the system (our constructions work also if we assume algebraic constraints see e.g. \cite{stinis2018}). Except for special cases, it is not advisable to try to enforce the differential equations directly at the continuum level. Instead we can discretize the equations in time using various numerical methods. We want to incorporate the discretized dynamics into the training process of the neural network. The purpose of such an attempt can be explained in two ways: (i) we want to aid the neural network so that it does not have to discover the dynamics (physics) from scratch; and (ii) we want the constraints to act as regularizers for the optimization problem which determines the parameters of the neural network.

Closer inspection of the concept of noisy data and of enforcing the discretized constraints reveals that they can be combined. However, this needs to be done with care. Recall that when we use noisy data we train the neural network to map a point from the noise cloud back to the noiseless point at the next time instant. Thus, we cannot enforce the discretized constraints as they are because the dynamics have been modified. In particular, the use of noisy data requires that the discretized constraints be modified to account {\it explicitly} for the restoring force. We have called the modification of the discretized constraints the {\it explicit} error-correction (see Section \ref{explicit_error_correction}). 

The meaning of the restoring force is analogous to that of memory terms in model reduction formalisms \cite{chorinstinis2007}. To see this, note that the flow map as well as the discretization of the original constraints are based on a {\it finite} timestep. The timescales that are smaller than the timestep used are {\it not} resolved explicitly. However, their effect on the resolved timescales cannot be ignored. In fact, it is what causes the inevitable error at each application of the flow map. The restoring force that we include in the modified constraints is there to remedy this error i.e. to account for the unresolved timescales albeit in a simplified manner. This is precisely the role played by memory terms in model reduction formalisms. In the current work we have restricted attention to {\it linear} error-correction terms. The linear terms come with coefficients whose magnitude is optimized as part of the training. In this respect, optimizing the error-correction term coefficients becomes akin to {\it temporal renormalization}. This means that the coefficients depend on the temporal scale at which we probe the system \cite{goldenfeld1992,barenblatt2003}. Finally, we note that the error-correction term can be more complex than linear. In fact, it can be modeled by a separate neural network. Results for such more elaborate error-correction terms will be presented elsewhere.  

We have implemented constraint enforcing in all three major modes of learning. For {\it supervised} learning, the constraints are added to the objective function (see Section \ref{supervised}). For the case of {\it unsupervised} learning, in particular generative adversarial networks \cite{goodfellowetal2014}, the constraints are introduced by augmenting the input of the discriminator (see Section \ref{unsupervised} and \cite{stinis2018}). Finally, for the case of {\it reinforcement} learning and in particular actor-critic methods \cite{sutton1999}, the constraints are added to the reward function. In addition, for the reinforcement learning case, we have developed a novel approach based on homotopy of the action-value function in order to stabilize and accelerate training (see Section \ref{reinforcement}).   

In recent years, there has been considerable interest in the development of methods that utilize data and physical constraints in order to train predictors for dynamical systems and differential equations e.g. see \cite{PhysRevE.91.032915, raissi2018, chen2018, Han8505, SIRIGNANO20181339, felsberger2018, wan2018, MaE9994} and references therein. Our approach is different, it introduces the novel concept of training on purpose with modified (noisy) data in order to incorporate (implicitly or explicitly) a restoring force in the dynamics learned by the neural network flow map. We have also provided the connection between the incorporation of such restoring forces and the concept of memory in model reduction. 

The paper is organized as follows. Section \ref{constraints_prediction} explains the need for constraints to increase the accuracy/efficiency of time series prediction as well as the form that these constraints can have. Section \ref{enforcing_constraints} presents ways to enforce such constraints in supervised learning (Section \ref{supervised}), unsupervised learning (Section \ref{unsupervised}) and reinforcement learning (Section \ref{reinforcement}). Section \ref{numerical} contains numerical results for the various constructions using the Lorenz system as an illustrative example. Finally, Section \ref{discussion} contains a brief discussion of the results as well as some ideas for current and future work.


\section{Constraints for time series prediction of dynamical systems}\label{constraints_prediction}

Suppose that we are given a dynamical system described by an M-dimensional set of differential equations 

\begin{equation}\label{odes}
\frac{dx}{dt}=f(x),
\end{equation} 
where $x \in \mathbb{R}^M.$ The system \eqref{odes} needs to be supplemented with an initial condition $x(0)=x_0.$ Furthermore, suppose that we are provided with time series data from the system \eqref{odes}. This means a sequence of points from a trajectory of the system $\{x_i^{data}\}_{i=1}^N$ recorded at time intervals of length $\Delta t.$ We would like to use this time series data to train a neural network to represent the flow map of the system i.e. a map $H^{\Delta t}$ with the property $H^{\Delta t}x(t)=x(t+\Delta t)$ for all $t$ and $x(t).$ 

We want to find ways to enforce {\it during training} the constraints implied by the system \eqref{odes}. Before we proceed, we should mention that in addition to \eqref{odes}, one could have extra constraints. For example, if the system \eqref{odes} is Hamiltonian, we have extra algebraic constraints since the system must evolve on an energy surface determined by its initial condition. We note that the framework we present below can also enforce algebraic constraints but in the current work we will focus on the enforcing of dynamic constraints like the system \eqref{odes}. Enforcing dynamic constraints is more demanding than enforcing algebraic ones. Technically, the enforcing of algebraic constraints requires only knowledge of the state of the system. On the other hand, the enforcing of dynamic constraints requires knowledge of the state {\it and} of the rate of change of the state. 

It is not advisable to attempt enforcing directly the constraints in \eqref{odes}. To do that requires that the output of the neural network includes both the state of the system and its rate of change. This {\it doubles} the dimension of the output and makes the necessary neural network size larger and its training more demanding. Instead, we will enforce constraints that involve only the state, albeit at more than one instants. For example, we can consider the simplest temporal discretization scheme, the forward Euler scheme \cite{haireretal1987}, and discretize \eqref{odes} with timestep $\Delta t$ to obtain

\begin{equation}\label{odes_discrete_1}
\hat{x}(t+\Delta t)=\hat{x}(t)+\Delta t f(\hat{x}(t))
\end{equation}  

Then, we can choose to enforce \eqref{odes_discrete_1} during training of the flow map. To be more precise, we can train the neural network representation of the flow map such that \eqref{odes_discrete_1} holds for all the training data $\{x_i^{data}\}_{i=1}^N.$ In addition, one can consider more elaborate temporal discretization schemes e.g. explicit Runge-Kutta methods \cite{haireretal1987}. In such a case, \eqref{odes_discrete_1} is replaced by
\begin{equation}\label{odes_discrete_2}
\hat{x}(t+\Delta t)=\hat{x}(t)+\Delta t f^{RK}(\hat{x}(t))
\end{equation}  
where $f^{RK}(\hat{x}(t))$ represents the functional form of the Runge-Kutta update. 

Such an approach of enforcing the constraint is {\it not} enough to guarantee that the trained neural network representation of the flow map will be accurate. In fact, as can be seen by simple numerical examples \cite{stinis2018}, the trained neural network flow map can lose its {\it predictive} ability rather fast. The reason is that we have used data from a time series i.e. a trajectory to train the neural network. However, a single trajectory is extremely unlikely (has measure zero) in the phase space of the system. Thus, the trained network predicts accurately {\it as long as the predicted state remains on the training trajectory}. But this is impossible, since the action of the flow map at every (finite) timestep involves an inevitable approximation error. If left unchecked, this approximation error causes the prediction to deviate into a region of phase space that the network has never trained on. Soon after, all the predictive ability of the network is lost.   

This observation highlights the need for an alternate way of enforcing the constraints. In fact, as we will explain now, it points towards the need for the enforcing of {\it alternate} constraints altogether. In particular, this observation underlines the need for the training of the neural network to include some kind of error-correcting mechanism. Such an error-correcting mechanism can help restore the trajectory predicted by the learned flow map when it inevitably starts deviating due to the finiteness of the used timestep. 

The way we have devised to implement this error-correcting mechanism can be {\it implicit} or {\it explicit}. By implicit we mean that we do not specify the functional form of the mechanism but only what we want it to achieve (Section \ref{implicit_error_correction}). On the other hand, the explicit implementation of the error-correcting mechanism does involve the specification of the {\it functional form} of the mechanism (Section \ref{explicit_error_correction}). 

The common ingredient for both implicit and explicit implementations is the use of a {\it noisy} version of the data during training. The main idea is the fact that the training of the neural network must address the inevitability of error that comes with the use of a finite timestep. For example, suppose that we are given data that are recorded every $\Delta t.$ The flow map we wish to train will produce states of the system at time instants that are $\Delta t$ apart. Every time the flow map is applied, even if it is applied on a point from the exact trajectory it will produce a state that has deviated from the exact trajectory at the next timestep. So, the trained flow map must learn how to correct such a deviation. 

\subsection{Implicit error-correction}\label{implicit_error_correction}
The {\it implicit} implementation of the error-correcting mechanism can be realized by the following simple procedure. We can consider each point of the given time series data and we can enhance it by a (random) cloud of points centered at the point on the time series. Such a cloud of points accounts for our ignorance about the inevitable error that the flow map commits at every step. The next step is to train the neural network to map a point from this cloud back to the {\it noiseless} trajectory at the next timestep. In this way, the neural network is trained to incorporate an error-correcting mechanism {\it implicitly}.  

Of course, there are the questions of the extent of the cloud of noisy points as well as the number of samples we need from it. These parameters depend on the magnitude of the interval $\Delta t$ and the accuracy of the training data. For example, if the training data were produced by a numerical method with a known order of accuracy then we expect the necessary extent of the noisy cloud to follow a scaling law with respect to the interval $\Delta t.$ Similarly, if the training data were produced by a numerical experiment with known measurement error, we expect the extent of the noisy cloud to depend on the measurement error.      

\subsection{Explicit error-correction}\label{explicit_error_correction}    
The {\it explicit} implementation of the error-correcting mechanism requires the specification of the functional form of the mechanism in addition to enhancing the given time series data by a noisy cloud. The main idea is that the need for the incorporation of the error-correcting mechanism means that the flow map we have to learn is not of the original system but of a {\it modified} system. Symbolically, the dynamics that the neural network based flow map must learn are given by ``learned dynamics = original dynamics + error-correction". As we have explained before, the error-correction term is needed due to the inevitable error caused by the use of a finite timestep. Such error-correction terms can be interpreted as memory terms appearing in model reduction formalisms \cite{chorinstinis2007}. However, note that here the reduction is in the {\it temporal} sense since it is caused by the use of a {\it finite} timestep. It can be thought of as a way to account for all the timescales that are contained in the interval $\Delta t$ and it is akin to {\it temporal} renormalization. Another way to interpret such an error-correction term is as a control mechanism \cite{isidori1995}.  

For the specific case of the forward Euler scheme given in \eqref{odes_discrete_1}, the explicit implementation of the error-correcting mechanism will mean that we want our trained flow map to satisfy 

\begin{equation}\label{odes_discrete_modified}
\hat{x}(t+\Delta t)=\hat{x}(t)+\Delta t f(\hat{x}(t)) + \Delta t f^{C}(\hat{x}(t)),
\end{equation}  
where $f^{C}(\hat{x}(t))$ is the error-correcting term. The obvious question is what is the form of $f^{C}(\hat{x}(t)).$ The simplest  state-dependent approximation is to assume $f^{C}(\hat{x}(t))$ is a linear function of the state $\hat{x}(t).$ For example, $f^{C}(\hat{x}(t))= A \hat{x}(t),$ where $A$ is a $M \times M$ matrix whose entries need to be determined. The entries of $A$ can be estimated during the training of the flow map neural network. There is no need to restrict the form of the correction term $f^{C}(\hat{x}(t))$ to a linear one. In fact, we can consider a {\it separate} neural network to represent $f^{C}(\hat{x}(t)).$ We have explored such constructions although a detailed presentation will appear in a future publication. A further question to be explored is the dependence of the elements of $A$ or of the parameters of the network for the representation of the error-correcting term on the timestep $\Delta t.$ In fact, we expect a scaling law dependence on $\Delta t$ which would be a manifestation of {\it incomplete similarity} \cite{barenblatt2003}. 

We also note that there is a further generalization of the error-correcting term $f^{C}(\hat{x}(t)),$ if we allow it to depend on the state of the system for times before $t.$ Given the analogy to memory terms alluded to above, such a dependence on the history of the evolution of the state of the system is an instance of a {\it non-Markovian} memory \cite{chorinstinis2007}. 

Finally, we note that \eqref{odes_discrete_modified} offers one more way to interpret the error-correction term, namely as a {\it modified} numerical method, a modified Euler scheme in this particular case, where the role of the error-correction term is to account for the error of the Euler scheme.


\section{Enforcing constraints in supervised, unsupervised and reinforcement learning}\label{enforcing_constraints}

In this section we will examine ways of enforcing the constraints in the 3 major modes of learning, namely supervised, unsupervised and reinforcement learning. 

\subsection{Supervised learning}\label{supervised}
The case of {\it supervised} learning is the most straightforward. Let us assume that the flow map is represented by a deep neural network denoted by $G$ depending on the parameter vector $\theta_G$ (the parameter vector $\theta_G$ contains the weights and biases of the neural network). The simplest objective function is the $L_2$ discrepancy between the network predictions and the training data trajectory given by
\begin{equation}\label{supervised_loss}
Loss_{supervised}=\frac{1}{\Lambda}\sum_{i=1}^{\Lambda} (G(z_{i})-x_i^{data})^2,
\end{equation}
where $z_i$ is a point from the noise cloud around a point of the given training trajectory and $x_i^{data}$ is the {\it noiseless} point on the given training trajectory after time $\Delta t.$ Note that we allowed freedom here in choosing the value of $\Lambda$ to accommodate various implementation choices e.g. number of samples from the noise cloud, mini-batch sampling etc. The parameter vector $\theta_G$ can be estimated by minimizing the objective function $Loss_{supervised}.$

For the sake of simplicity, suppose that we want to enforce the constraints given in  \eqref{odes_discrete_modified} with a {\it diagonal} linear representation of the error-correcting term i.e. $f_j^{C}(\hat{x}(t))= -a_j \hat{x}_j(t),$ for $j=1,\ldots,M$ (the minus sign is in anticipation of this being a restoring force). Then we can consider the modified objective function given by
\begin{gather}Loss_{supervised}^{constraints}=\frac{1}{\Lambda}\bigg[ \sum_{i=1}^{\Lambda} (G(z_{i})-x_i^{data})^2 \notag \\
+ \sum_{j=1}^M (G_j(z_{i})-z_{ij}-\Delta t f_j(z_{ij}) + \Delta t a_j z_{ij})^2 \bigg], \label{supervised_loss_constraints}
\end{gather}
where $z_{ij}$ is the $j$-th component of the noise cloud point $z_i$ and $f_j$ is the $j$-th component of the vector $f.$ Notice that the minimization of the modified objective function $Loss_{supervised}^{constraints}$ leads to the determination of both the parameter vector $\theta_G$ {\it and} the error-correcting representation parameters $a_j, \; j=1,\ldots,M.$ Also note, that if instead of the forward Euler scheme we use e.g. a more elaborate Runge-Kutta method as given in \eqref{odes_discrete_2}, then we can still use \eqref{supervised_loss_constraints} but with the vector $f$ replaced by $f^{RK}.$


\subsection{Unsupervised learning - Generative Adversarial Networks}\label{unsupervised}

The next mode of learning that we will examine how to enforce constraints for is {\it unsupervised} learning, in particular Generative Adversarial Networks (GANs) \cite{goodfellowetal2014}. This material appeared first in \cite{stinis2018} albeit with different notation. We repeat it here with the current notation for the sake of completeness.

Generative Adversarial Networks comprise of two networks, a generator and a discriminator. The target is to train the generator's output distribution $p_g(x)$ to be close to that of the true data $p_{data}.$ We define a prior input $p_z(z)$ on the generator input variables $z$ and a mapping $G(z;\theta_G)$ to the data space where $G$ is a differentiable function represented by a neural network with parameters $\theta_G.$ We also define a second neural network (the discriminator) $D(x;\theta_D),$ which outputs the probability that $x$ came from the true data distribution $p_{data}$ rather than $p_g.$ We train $D$ to {\it maximize} the probability of assigning the correct label to both training examples and samples from the generator $G.$ Simultaneously, we train $G$ to minimize $\log (1-D(G(z))).$ We can express the adversarial nature of the relation between $D$ and $G$ as the two-player min-max game with value function $V(D,G)$:
\begin{equation}\label{game_gan}
\min_G \max_D V(D,G) = E_{x \sim p_{data}(x)}[\log D(x)] +E_{z \sim p_z(z)}[\log (1- D(G(z)))].
\end{equation}
The min-max problem can be formulated as a bilevel minimization problem for the discriminator and the generator using the objective functions $-E_{x \sim p_{data}(x)}[\log D(x)]$ and $-E_{t \sim p_z(z)}[\log (D(G(z)))]$ respectively. The modification of the objective function for the generator has been suggested to avoid early saturation of $\log (1- D(G(z)))$ due to the faster training of the discriminator \cite{goodfellowetal2014}. On the other hand, while this modification avoids saturation, the well-documented instability of GAN training appears \cite{arjovskybottou2017}. Even though the min-max game can be formulated as a bilevel minimization problem, in practice the discriminator and generator neural networks are usually updated iteratively.

We are interested in training the generator $G$ to represent the flow map of the dynamical system. That means that if $z$ is the state of the system at a time instant $t,$ we would like to train the generator $G$ to produce as output $G(z),$ an accurate estimate of the state of the system at time $t+ \Delta t.$ In \cite{stinis2018} we have presented a way to enforce constraints in the output of the generator $G$ that respects the game-theoretic setup of GANs. We can do so by {\it augmenting} the input of the discriminator by the constraint residuals i.e. how well does a sample satisfy the constraints. Of course, such an augmentation of the discriminator input should be applied {\it both} to the generator-created samples as well as the samples from the true distribution. This means that we consider a two-player min-max game with the modified value function 
\begin{gather}
\min_G \max_D V^{constraints}(D,G) \notag \\
= E_{x \sim p_{data}(x)}[\log D(x,\epsilon_D(x))] +E_{z \sim p_z(z)}[\log (1- D(G(z),\epsilon_G(z)))],\label{game_gan_constraints}
\end{gather}     
where $\epsilon_D(x)$ is the constraint residual for the true sample and $\epsilon_G(z)$ is the constraint residual for the generator-created sample. Note that in our setup, the generator input distribution $p_z(z)$ will be from the noise cloud around the training trajectory. On the other hand, the true data distribution $p_{data}$ is the distribution of values of the (noiseless) training trajectory.

As explained in \cite{stinis2018}, taking the constraint residual $\epsilon_D(x)$ to be zero for the true samples can exacerbate the well-known saturation (instability) issue with training GANs. Thus, we take $\epsilon_D(x)$ to be a random variable with mean zero and small variance dictated by Monte-Carlo or other numerical/mathematical/physical considerations.  On the other hand, for $\epsilon_G(z)$ we can use the constraint we want to enforce. For example, for the constraint based on the forward Euler scheme with {\it diagonal} linear error-correcting term, we take $\epsilon^j_G(z)=G_j(z)-z_{j}-\Delta t f_j(z_{j}) + \Delta t a_j z_{j}$ for $j=1,\ldots,M,$ where $z$ is a sample from the noise cloud around a point of the training time series data. The expression for the constraint residual $\epsilon^j_G(z)$ can be easily generalized for more elaborate numerical methods and error-correcting terms.


\subsection{Reinforcement learning - Actor-critic methods}\label{reinforcement}

The third and final mode of learning for which we will examine how to enforce constraints for time series prediction is {\it reinforcement} learning and in particular Actor-Critic (AC) methods \cite{sutton1999,grondman2012,silver2014,lillicrap2015,pfau2016}. We will also present a novel approach based on {\it homotopy} in order to stabilize and accelerate training.

\subsubsection{General setup of AC methods}

We will begin with the general setup of an AC method and then provide the necessary modifications to turn it into a computational device for training flow map neural network representations. The setup consists of an agent ({\it actor}) interacting with an environment in discrete timesteps. At each timestep $t$ the agent is supplied with an observation of the environment and the agent state $s_t.$ Based on the state $s_t$ it takes an action $a_t$ and receives a scalar reward $r_t.$ An agent's behavior is based on a action policy $\pi,$ which is a map from the states to a probability distribution over the actions, $\pi: \; \mathcal{S} \rightarrow \mathcal{P_{\pi}}(\mathcal{A})$ where $\mathcal{S}$ is the state space and $\mathcal{A}$ is the action space. We also need to specify an initial state distribution $p_0(s_0),$ the transition function $\mathcal{P}(s_{t+1}|s_t,a_t)$ and the reward distribution $\mathcal{R}(s_t,a_t).$  


The aim of an AC method is to learn in tandem an action-value function ({\it critic})
\begin{equation}\label{action_value}
Q^{\pi}(s_t,a_t)=\mathbb{E}_{s_{t+k+1} \sim \mathcal{P}, r_{t+k} \sim \mathcal{R}, a_{t+k+1} \sim \mathcal{\pi}}\bigg[ \sum_{k=0}^{\infty} \gamma^k r_{t+k} \bigg| s_t,a_t\bigg]
\end{equation}
and an action policy that is optimal for the action-value function
\begin{equation}\label{policy}
\pi^* = \arg \underset{\pi}{\max} \; \mathbb{E}_{s_0 \sim p_0,a_0 \sim \pi}[Q^{\pi}(s_0,a_0)].
\end{equation}
The parameter $\gamma \in [0,1]$ is called the {\it discount factor} and it expresses the degree of trust in future actions. Eq. \eqref{action_value} can be rewritten in a recursive manner as
\begin{equation}\label{bellman}
Q^{\pi}(s_t,a_t)=\mathbb{E}_{r_t \sim \mathcal{R}, s_{t+1} \sim \mathcal{P}}[r_t+ \gamma \mathbb{E}_{a_{t+1}\sim \pi}[Q^{\pi}(s_{t+1},a_{t+1})]]
\end{equation}
which is called the Bellman equation. Thus, the task of finding the action-value function is equivalent to solving the Bellman equation. We can solve the Bellman equation by reformulating it as an optimization problem
\begin{equation}\label{bellman_opt}
Q^{\pi}= \arg \underset{Q}{\min} \mathbb{E}_{s_t,a_t \sim \pi}\big[ (Q(s_t,a_t)-y_t)^2\big] 
\end{equation}
where 
\begin{equation}\label{bellman_opt_target}
y_t=\mathbb{E}_{r_t \sim \mathcal{R}, s_{t+1} \sim \mathcal{P}, a_{t+1}\sim \pi}[r_t+ \gamma Q(s_{t+1},a_{t+1})]
\end{equation}
is called the {\it target}. In \eqref{bellman_opt}, instead of the square of the distance of the action-value function from the target, we could have used any other divergence that is positive except when the action-value function and target are equal \cite{pfau2016}. Using the objective functions $\mathbb{E}_{s_t,a_t \sim \pi}\big[ (Q(s_t,a_t)-y_t)^2\big] $ and $-\mathbb{E}_{s_0 \sim p_0,a_0 \sim \pi}[Q^{\pi}(s_0,a_0)]$ for the action-value function and action policy respectively, we can express the task of reinforcement learning also as a bilevel minimization problem \cite{pfau2016}. However, as in the case of GANs discussed before, in practice the action-value function and action policy are usually updated iteratively.

Before we adapt the AC setup to our task of enforcing constraints for time series prediction we will focus on two special choices: (i) the use of {\it deterministic} target policies and (ii) the of use neural networks to represent both the action-value function and the action policy \cite{silver2014,lillicrap2015}. 

We start with the effect of using a deterministic target policy denoted as $\mu: \mathcal{S} \rightarrow \mathcal{A}.$ Then, the Bellman equation \eqref{bellman} can be written as
\begin{equation}\label{bellman_deterministic}
Q^{\mu}(s_t,a_t)=\mathbb{E}_{r_t \sim \mathcal{R}, s_{t+1} \sim \mathcal{P}}[r_t+ \gamma Q^{\mu}(s_{t+1},\mu(s_{t+1}))]
\end{equation}
Note that the use of a deterministic target policy $a_{t+1}$ has allowed us to drop the expectation with respect to $a_{t+1}$ that appeared in \eqref{bellman_opt_target} and find
\begin{equation}\label{bellman_opt_target_drop}
y_t=\mathbb{E}_{r_t \sim \mathcal{R}, s_{t+1} \sim \mathcal{P}}[r_t+ \gamma Q(s_{t+1},\mu(s_{t+1})].
\end{equation}
Also, note that the expectations in \eqref{bellman_deterministic} and \eqref{bellman_opt_target_drop} depend only on the environment. This means that it is possible to learn $Q^{\mu}$ off-policy, using transitions that are generated from a different stochastic behavior policy $\beta$. We can rewrite the optimization problem \eqref{bellman_opt}-\eqref{bellman_opt_target} as
\begin{equation}\label{bellman_opt_off}
Q^{\mu}= \arg \underset{Q}{\min} \mathbb{E}_{s_t \sim \rho^{\beta},a_t \sim \beta, r_t \sim \mathcal{R}}\big[ (Q(s_t,a_t)-y_t)^2\big] 
\end{equation}
where 
\begin{equation}\label{bellman_opt_target_off}
y_t=r_t+ \gamma Q(s_{t+1},\mu(s_{t+1})).
\end{equation}
The state visitation distribution $\rho^{\beta}$ is related to the policy $\beta.$ We will use below this flexibility to introduce our noise cloud around the training trajectory.

We continue with the effect of using neural networks to represent both the action-value function and the policy. We restrict attention to the case of a deterministic policy since this will be the type of policy we will use later for our time series prediction application. To motivate the introduction of neural networks we begin with the concept of $Q$-learning as a way to learn the action-value function and the policy \cite{watkins1992,mnih2015}. In $Q$-learning, the optimization problem \eqref{bellman_opt_off}-\eqref{bellman_opt_target_off} to find the action-value function is coupled with the greedy policy estimate $\mu(s)=\arg \underset{a}{\max}\;Q(s,a).$ Thus, the greedy policy requires an optimization at every timestep. This can become prohibitively costly for the type of action spaces that are encountered in many applications. This has led to (i) the adoption of (deep) neural networks for the representation of the action-value function and the policy and (ii) the update of the neural network for the policy after {\it each} $Q$-learning iteration for the action-value function \cite{silver2014}.

We assume that the action-value function $Q(s_t,a_t|\theta_Q)$ is represented by a neural network with parameter vector $\theta_Q$ and the deterministic policy $\mu(s|\theta_{\mu})$ by a neural network with parameter vector $\theta_{\mu}.$ The deterministic policy gradient algorithm \cite{silver2014} uses \eqref{bellman_opt_off}-\eqref{bellman_opt_target_off} to learn $Q(s_t,a_t|\theta_Q)$. The policy $\mu(s|\theta_{\mu})$ is updated after every iteration of the $Q$-optimization using the policy gradient
\begin{gather}
\nabla_{\theta_{\mu}} \mathbb{E}_{s_t \sim \rho^{\beta}}[Q(s,a|\theta_Q)|_{s=s_t,a=\mu(s_t|\theta_{\mu})}] \notag \\
= \mathbb{E}_{s_t \sim \rho^{\beta}}[\nabla_{\theta_{\mu}}Q(s,a|\theta_Q)|_{s=s_t,a=\mu(s_t|\theta_{\mu})}]  \label{policy_gradient}
\end{gather}    
which can be computed through the chain rule \cite{silver2014,lillicrap2015}.

\subsubsection{AC methods for time series prediction and enforcing constraints}\label{enforcing_ac}

We explain now how an AC method can be used to train the flow map of a dynamical system. In addition, we provide a way for enforcing constraints during training. 

We begin by identifying the state $s_t$ with the state of the dynamical system at time $t.$ Also, we identify the discrete timesteps with the iterations of the flow map that advance the state of the dynamical system by $\Delta t$ units in time. The action policy $\mu(s_t|\theta_{\mu})$ is the action that needs to be taken to bring the state of the system from $s_t$ to $s_{t+1}.$ However, instead of learning separately the action policy that results in $s_t$ being mapped to $s_{t+1},$ we can {\it identify} the policy $\mu(s_t|\theta_{\mu})$ with the state $s_{t+1}$ i.e. $\mu(s_t|\theta_{\mu})=s_{t+1}.$ In this way, training for the policy $\mu(s_t|\theta_{\mu})$ is equivalent to training for the flow map of the dynamical system. 

We also take advantage of the off-policy aspect of \eqref{bellman_opt_off}-\eqref{bellman_opt_target_off} to choose the distribution of states $\rho^{\beta}$ to be the one corresponding to the noise cloud around the training trajectory needed to implement the error-correction. Thus, we see that the intrinsic statistical nature of the AC method bodes well with our approach to error-correcting. 

To complete the specification of the AC method as a method for training the flow map of a dynamical system we need to specify the reward function. The specification of the reward function is an important aspect of AC methods and reinforcement learning in general \cite{sorg2010,Guo2016}. We have chosen a simple {\it negative} reward function. To conform with the notation from Sections \ref{supervised} and \ref{unsupervised} we specify the reward function as
\begin{equation}\label{reward}
r(z,x)=-\sum_{j=1}^{M} (\mu_j(z)-x_j^{data})^2, 
\end{equation}
where $z$ is a point on the noise cloud at the time $t,$ $\mu_j(z)$ is the $j$-th component of its image through the flow map and $x_j^{data}$ is the $j$-th component of the {\it noiseless} point on the training trajectory at time $t+\Delta t$ that it is mapped to. Similarly, for the case when we want to enforce constraints e.g. {\it diagonal} linear representation for the error-correcting term we can define the reward function as  
\begin{equation}\label{reward_constraints}
r(z,x)=-\sum_{j=1}^{M} \bigg[ (\mu_j(z)-x_j^{data})^2 + (\mu_j(z)-z_j-\Delta t f_j(z) + \Delta t a_j z_j)^2 \bigg].
\end{equation}
For each time $t,$ the reward function that we have chosen uses information only from the state of the system at time $t$ and $t+\Delta t.$ Of course, how much credit we assign to this information is determined by the value of the discount factor $\gamma.$ 

If $\gamma=0,$ then we disregard any information beyond time $t+\Delta t.$ In this case, the AC method becomes a supervised learning method in disguise. In fact, from \eqref{bellman_opt_off}-\eqref{policy_gradient} we see that when $\gamma=0,$ the task of maximizing the action-value function is equivalent to maximizing the reward. The maximum value for our {\it negative} reward \eqref{reward} is 0 which is the optimal value for the supervised learning loss function \eqref{supervised_loss} (the average over the noise cloud in \eqref{supervised_loss} is the same as the average w.r.t. to $\rho^{\beta}$). A similar conclusion holds for the case of the reward with constraints \eqref{reward_constraints} and the supervised learning loss function with constraints \eqref{supervised_loss_constraints}.  

If on the other hand we set $\gamma=1,$ we assign equal importance to current and future rewards. This corresponds to the case where the environment is deterministic and thus the same actions result in the same rewards.    

\subsubsection{Homotopy for the action-value function}\label{homotopy_action}

AC methods utilizing deep neural networks to represent the action-value function and the policy have proven to be difficult to train due to instabilities. As a result, a lot of techniques have been devised to stabilize training (see e.g. \cite{pfau2016} and references therein for a review of stabilizing techniques). In our numerical experiments we tried some of these techniques but could not get satisfactory training results neither for the action-value function nor for the policy. This is the reason we developed a novel approach based on homotopy which indeed resulted in successful training (see results in Section \ref{numerical_reinforcement}).  

To motivate our approach we examine the case when $\gamma=0,$ although similar arguments hold for the other values of $\gamma.$ As we have discussed in the previous section, when $\gamma=0,$ the AC method is a supervised learning method in disguise. In fact, the AC method tries to achieve the same result as a supervised learning method but does it in a rather inefficient way. If we look at \eqref{bellman_opt_off}-\eqref{policy_gradient}, we see that in the action-value update (which is effected through \eqref{bellman_opt_off}), the AC method tries to minimize the distance between the action-value function and the reward function. Then, in the action policy update step (which is effected through the use of \eqref{policy_gradient}), the AC method tries to maximize the action-value function. In essence, through this two-step procedure, the AC method tries to maximize the reward but does so in a roundabout way. 

If we think of a plane (in function space) where on one axis we have the action-value function $Q(s_t,a_t)$ and on the other the reward $r_t$, we are trying to find the point on the line $Q(s_t,a_t)=r_t$ which maximizes $Q(s_t,a_t).$ But hitting this line from a random initialization of the neural networks for $Q(s_t,a_t)$ and of the policy $\mu(s_t)$ is extremely unlikely. We would be better off if we started our optimization from a point {\it on} the line and then look for the maximum of $Q(s_t,a_t).$ In other words, for the case of $\gamma=0,$ we have a better chance of training accurately if we let $Q(s_t,a_t)=r_t$ in \eqref{policy_gradient}. A similar argument for the case $\gamma \neq 0$ shows why we will have a better chance of training if we let $Q(s_t,a_t)=r_t + \gamma Q(s_{t+1},\mu(s_{t+1}))$ in \eqref{policy_gradient}. 

There is an extra mathematical reason why the identification $Q(s_t,a_t)=r_t + \gamma Q(s_{t+1},\mu(s_{t+1}))$ can result in better training. Recall from \eqref{reward_constraints} that the reward function $r_t$ contains {\it all} the information from the training trajectory and the constraints we wish to enforce. In addition, $r_t$ depends on the parameter vector $\theta_{\mu}$ for the neural network that represents the action policy $\mu.$ Thus, when we use the expression $r_t + \gamma Q(s_{t+1},\mu(s_{t+1}))$ in \eqref{policy_gradient} for the update step of $\theta_{\mu},$ we back-propagate {\it directly} the available information from the training trajectory and the constraints to $\theta_{\mu}.$ This is because we differentiate directly $r_t$ w.r.t. $\theta_{\mu}.$ On the other hand, in the original formulation we do {\it not} differentiate $r_t$ at all, because there $r_t$ appears {\it only} in the update step for the action-value function. That update step involves differentiation w.r.t. the action-value function parameter vector $\theta_{Q}$ but not $\theta_{\mu}.$   

Of course, if we make the identification $Q(s_t,a_t)=r_t + \gamma Q(s_{t+1},\mu(s_{t+1}))$ in \eqref{policy_gradient} we have modified the original problem. The question is how is the solution to the modified problem related to the original one. Through algebraic inequalities, one can show that the optimum for $Q(s_t,a_t)$ for the modified problem provides a lower bound on the optimum for the original problem. It can also provide an upper bound if we make extra assumptions about the difference $Q(s_t,a_t)-r_t - \gamma Q(s_{t+1},\mu(s_{t+1}))$ e.g. the convex-concave assumptions appearing in the min-max theorem \cite{osborne2004}.   

To avoid the need for such extra assumptions, we have developed an alternative approach. We initialize the training procedure with the identification $Q(s_t,a_t)=r_t + \gamma Q(s_{t+1},\mu(s_{t+1}))$ in \eqref{policy_gradient}. As the training progresses we morph the modified problem back to the original one via {\it homotopy}. In particular, we use in \eqref{policy_gradient} instead of $Q(s_t,a_t)$ the expression 
\begin{equation}
\delta \times Q(s_t,a_t) + (1-\delta) \times [r_t + \gamma Q(s_{t+1},\mu(s_{t+1}))],
\end{equation}       
where $\delta$ is the homotopy parameter. A user-defined schedule evolves $\delta$ during training from 0 (modified problem) to 1 (original problem). The accuracy of the training is of course dependent on the schedule for $\delta.$ However, in our numerical experiments we obtained good results without the need for a very refined schedule. One general rule of thumb is that the schedule should be slower for larger values of $\gamma$ i.e. allow more iterations between increases in the value of $\delta.$ This is to be expected because for larger values of $\gamma,$ the influence of $r_t$ in the optimization of $r_t + \gamma Q(s_{t+1},\mu(s_{t+1}))$ is reduced. Thus, it is more difficult to back-propagate the information from $r_t$ to the action policy parameter vector $\theta_{\mu}.$ However, note that larger values of $\gamma$ allow us to take more into account future rewards, thus allowing the AC method to be more versatile.


\section{Numerical results}\label{numerical}

We use the example of the Lorenz system to illustrate the constructions presented in Sections \ref{constraints_prediction} and \ref{enforcing_constraints}.

The Lorenz system is given by 
\begin{align}
\frac{d x_1}{dt}&=\sigma (x_2-x_1)  \label{lorenz1} \\
\frac{d x_2}{dt}&= \rho x_1 - x_2 - x_1 x_3  \label{lorenz2}  \\
\frac{d x_3}{dt}&= x_1 x_2 - \beta x_3 \ \label{lorenz3}
\end{align}
where $\sigma, \rho$ and $\beta$ are positive. We have chosen for the numerical experiments the commonly used values $\sigma=10,$ $\rho=28$ and $\beta=8/3.$ For these values of the parameters the Lorenz system is chaotic and possesses an attractor for almost all initial points. We have chosen the initial condition $x_1(0)=0,$ $x_2(0)=1$ and $x_3(0)=0.$  

We have used as training data the trajectory that starts from the specified initial condition and is computed by the Euler scheme with timestep $\delta t=10^{-4}.$ In particular, we have used data from a trajectory for $t \in [0,3].$ For all three modes of learning, we have trained the neural network to represent the flow map with timestep $\Delta t=1.5 \times 10^{-2}$ i.e. 150 times larger than the timestep used to produce the training data. After we trained the neural network that represents the flow map, we used it to predict the solution for $t \in [0,9].$ Thus, the trained flow map's task is to predict (through iterative application) the whole training trajectory for $t \in [0,3]$ starting from the given initial condition and then keep producing predictions for $t \in (3,9].$    

This is a severe test of the learned flow map's predictive abilities for four reasons. First, due to the chaotic nature of the Lorenz system there is no guarantee that the flow map can correct its errors so that it can follow closely the training trajectory even for the interval $[0,3]$ used for training. Second, by extending the interval of prediction beyond the one used for training we want to check whether the neural network has actually learned the map of the Lorenz system and not just overfitting the training data. Third, we have chosen an initial condition that is far away from the attractor but our integration interval is long enough so that the system does reach the attractor and then evolves on it. In other words, we want the neural network to learn both the evolution of the transient and the evolution on the attractor. Fourth, we have chosen to train the neural network to represent the flow map corresponding to a much larger timestep than the one used to produce the training trajectory in order to check the ability of the error-correcting term to account for a significant range of unresolved timescales (relative to the training trajectory).   

We performed experiments with different values for the various parameters that enter in our constructions. We present here indicative results for the case of $N=2\times10^4$ samples ($N/3$ for training, $N/3$ for validation and $N/3$ for testing). We have chosen $N_{cloud}=100$ for the cloud of points around each input. Thus, the timestep $\Delta t=1.5 \times 10^{-2}.$ This is because there are $20000/100=200$ time instants in the interval $[0,3]$ at a distance $\Delta t = 3/200=1.5 \times 10^{-2}$ apart. 

The noise cloud for the neural network at a point $t$ was constructed using the point $x_i(t)$ for $i=1,2,3,$ on the training trajectory and adding random disturbances so that it becomes the collection $x_{il}(t) (1-R_{range}+2R_{range} \times \xi_{il})$ where $l=1,\ldots,N_{cloud}.$ The random variables $\xi_{il} \sim U[0,1]$ and $R_{range} =2\times 10^{-2}.$ As we have explained before, we want to train the neural network to map the input from the noise cloud at a time $t$ to the  {\it noiseless} point $x_i(t + \Delta t)$ (for $i=1,2,3,$) on the training trajectory at time $t+ \Delta t.$

We have to also motivate the value of $R_{range}$ for the range of the noise cloud. Recall that the training trajectory was computed with the Euler scheme which is a first-order scheme. For the interval $\Delta t=1.5 \times 10^{-2}$ we expect the error committed by the flow map to be of similar magnitude and thus we should accommodate this error by considering a cloud of points within this range. We found that taking $R_{range}$ slightly larger and equal to $2\times 10^{-2}$ helps the accuracy of the training.

We denote by $(F_1(z),F_2(z),F_3(z))$ the output of the neural network flow map for an input $z.$ This corresponds to $(G_1(z),G_2(z),G_3(z))$ for the notation of Section \ref{supervised} (supervised learning) and \ref{unsupervised} (unsupervised learning) and to $(\mu_1(z),\mu_2(z),\mu_3(z))$ for the notation of Section \ref{reinforcement}. 

As explained in detail in \cite{stinis2018}, we employ a learning rate schedule that we have developed and which uses the relative error of the neural network flow map. For a mini-batch of size $m,$ we define the relative error as
\begin{gather*}
RE_m= \frac{1}{m}\sum_{j=1}^m \frac{1}{3} \biggl[ \frac{|F_1(z_j)-x_1(t_j+ \Delta t)|}{|x_1(t_j + \Delta t)|} + \frac{|F_2(z_j)-x_2(t_j+ \Delta t)|}{|x_2(t_j + \Delta t)|} \\
+\frac{|F_3(z_j)-x_3(t_j+ \Delta t)|}{|x_3(t_j + \Delta t)|}  \biggr] ,
\end{gather*}
where $(F_1(z_j),F_2(z_j),F_3(z_j))$ is the neural network flow map prediction at $t_j + \Delta t$ for the input vector $z_j=(z_{j1},z_{j2},z_{j3})$ from the noise cloud at time $t_j.$ Also, $(x_1(t_j + \Delta t),x_2(t_j + \Delta t),x_3(t_j + \Delta t))$ is the point on the training trajectory computed by the Euler scheme with $\delta t=10^{-4}.$ The tolerance for the relative error was set to $TOL = 1/\sqrt{N/3}=1/\sqrt{20^4/3} \approx 0.0122.$ (see \cite{stinis2018} for more details about $TOL$). For the mini-batch size we have chosen $m=1000$ for the supervised and unsupervised cases and $m=33$ for the reinforcement learning case.

We also need to specify the constraints that we want to enforce. Using the notation introduced above, we want to train the neural network flow map so that its output $(F_1(z_j),F_2(z_j),F_3(z_j))$ for an input data point $z_j=(z_{j1},z_{j2},z_{j3})$ from the noise cloud satisfies
\begin{align}
F_1(z_j) &= z_{j1} + \Delta t [\sigma (z_{j2}-z_{j1})] - \Delta t a_1  z_{j1}  \label{lorenz_modified1} \\
F_2(z_j) &= z_{j2} + \Delta t [\rho z_{j1} - z_{j2}- z_{j1} z_{j3}]-  \Delta t a_2  z_{j2}  \label{lorenz_modified2}  \\
F_3(z_j) &= z_{j3} + \Delta t [ z_{j1} z_{j2} - \beta z_{j3}] -  \Delta t a_3  z_{j3} \label{lorenz_modified3}
\end{align} 
where $a_1, a_2$ and $a_3$ are parameters to be optimized during training. The first two terms on the RHS of \eqref{lorenz_modified1}-\eqref{lorenz_modified3} come from the forward Euler scheme, while the third is the {\it diagonal} linear error-correcting term.

\subsection{Supervised learning}\label{numerical_supervised}

We begin the presentation of results with the case of {\it supervised} learning. Our aim in this subsection is threefold: (i) show that the {\it explicit} enforcing of the constraints is better than the {\it implicit} one, (ii) show that the addition of noise to the training trajectory is beneficial and (iii) show that the addition of error-correcting terms to the constraints can be beneficial {\it even} if we use the {\it noiseless} trajectory. The latter point highlights once again the promising influx to predictive machine learning of ideas from model reduction.    

\subsubsection{Implicit versus explicit constraint enforcing}

\begin{figure}[ht]
   \centering
   \subfigure[]{%
   \includegraphics[width = 5cm]{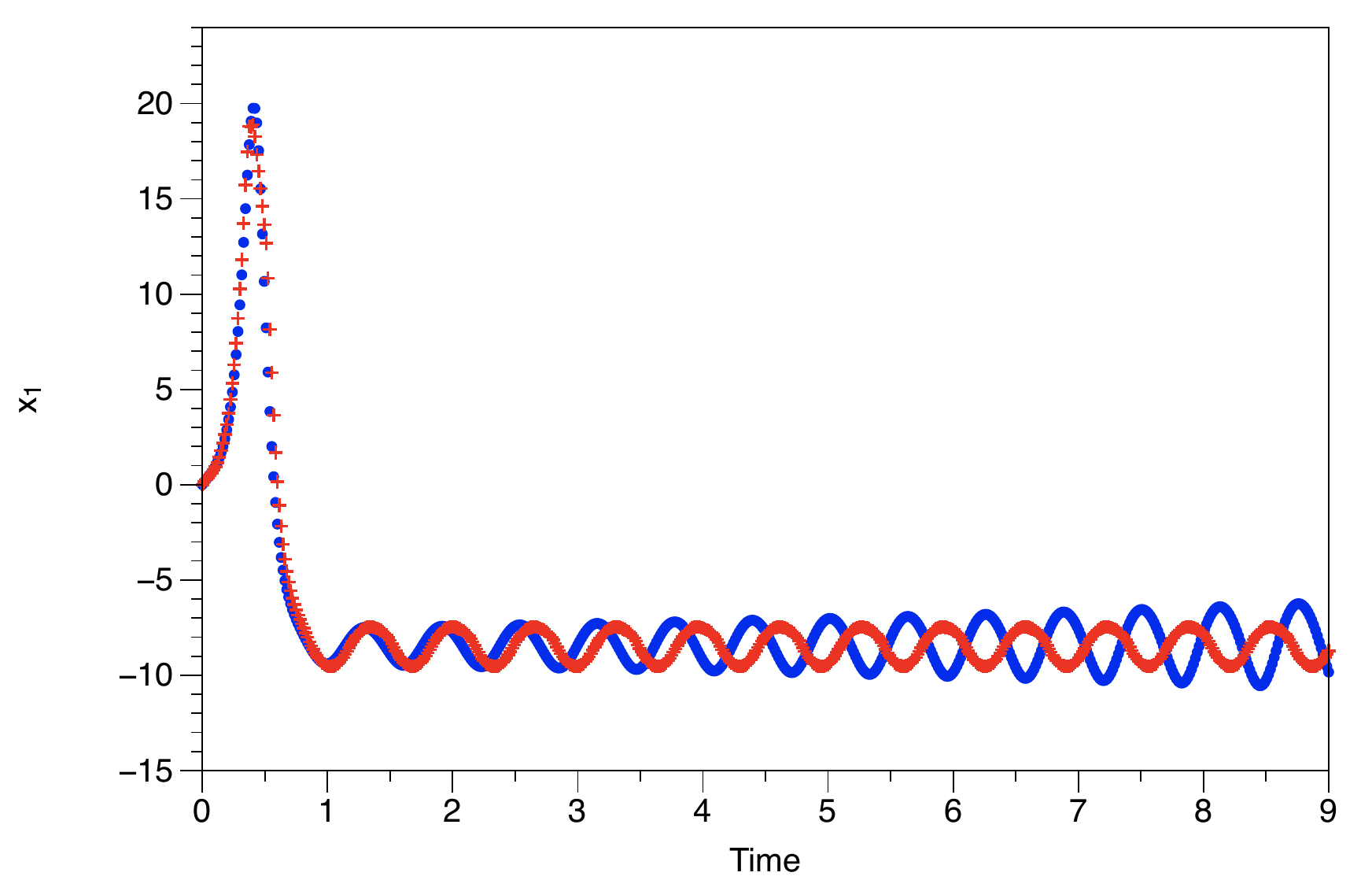}
   \label{plot_lorenz_superviseda}}
      \quad
   \subfigure[]{%
   \includegraphics[width = 5cm]{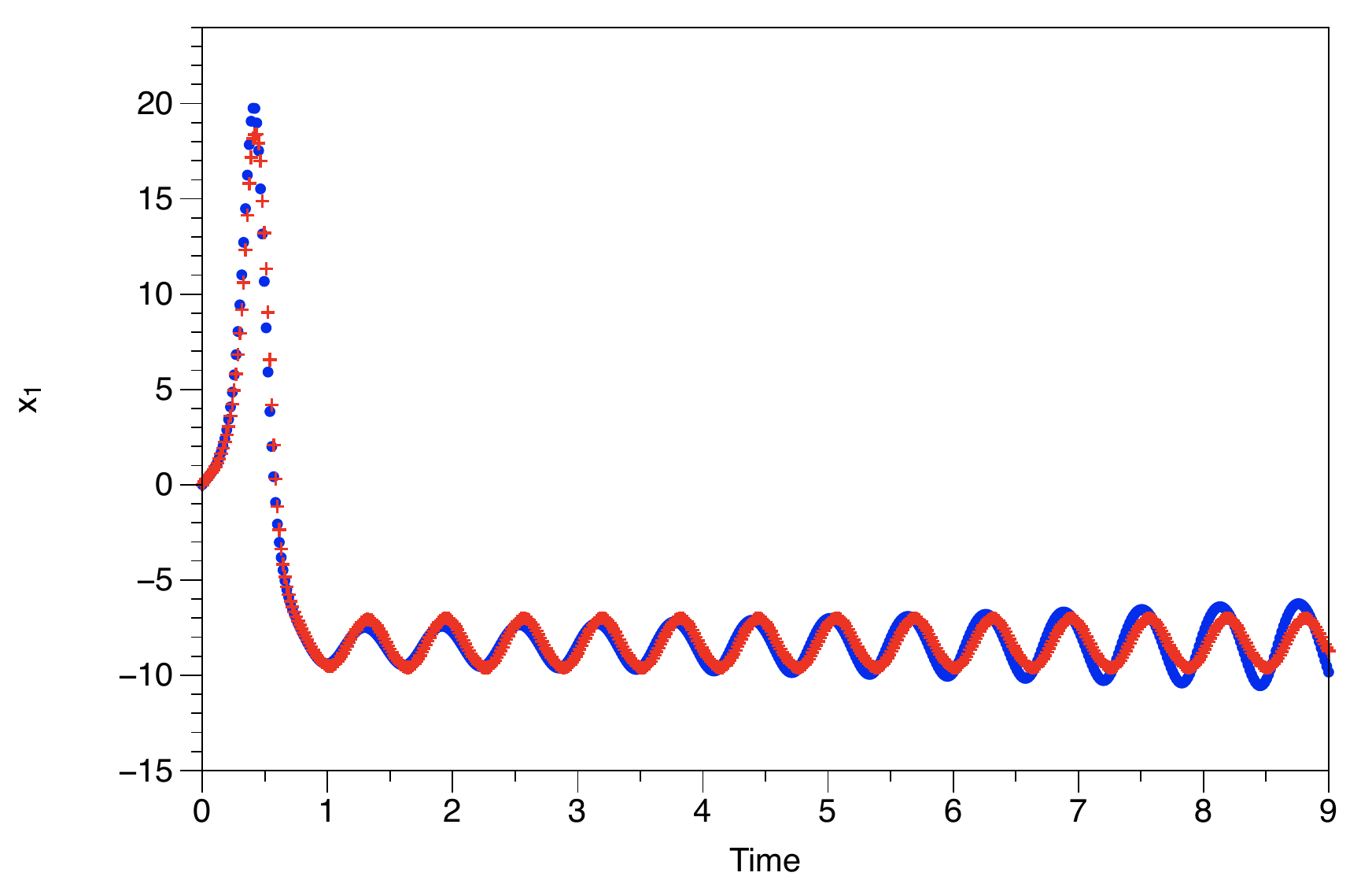}
   \label{plot_lorenz_supervisedb}}
\caption{Supervised learning. Comparison of ground truth for $x_1(t)$ computed with the Euler scheme with timestep $\delta t=10^{-4}$ (blue dots) and the neural network flow map prediction with timestep $\Delta t=1.5\times10^{-2}$ (red crosses). (a) {\it noisy} data {\it without enforced constraints} during training; (b) {\it noisy} data {\it with enforced constraints} during training (see text for details).  }
\label{plot_lorenz_supervised}
\end{figure}

We used a deep neural network for the representation of the flow map with 10 hidden layers of width 20. We note that because the solution of the Lorenz system acquires values outside of the region of the activation function we have removed the activation function from the last layer of the generator (alternatively we could have used batch normalization and kept the activation function). Fig. \ref{plot_lorenz_supervised} compares the evolution of the prediction for $x_1(t)$ of the neural network flow map starting at $t=0$ and computed with a timestep $\Delta t=1.5\times10^{-2}$ to the ground truth (training trajectory) computed with the forward Euler scheme with timestep $\delta t=10^{-4}.$ We show plots only for $x_1(t)$ since the results are similar for the $x_2(t)$ and $x_3(t).$ We want to make two observations. 

First, the prediction of the neural network flow map is able to follow with adequate accuracy the ground truth not only during the interval $[0,3]$ that was used for training, but also during the interval $(3,9].$ Second, the {\it explicit} enforcing of constraints i.e. the enforcing of the constraints \eqref{lorenz_modified1}-\eqref{lorenz_modified3} (see results in Fig. \ref{plot_lorenz_supervisedb}) is better than the {\it implicit} enforcing of constraints.    

\subsubsection{Noisy versus noiseless training trajectory}

We have advocated the use of a noisy version of the training trajectory in order for the neural network flow map to be exposed to larger parts of the phase space. The objective of such an exposure is to train the flow map to know how to respond to points away from the training trajectory where it is bound to wander due to the inevitable error committed through its repeated application during prediction. In this subsection we present results which corroborate our hypothesis.

\begin{figure}[ht]
   \centering
   \subfigure[]{%
   \includegraphics[width = 5cm]{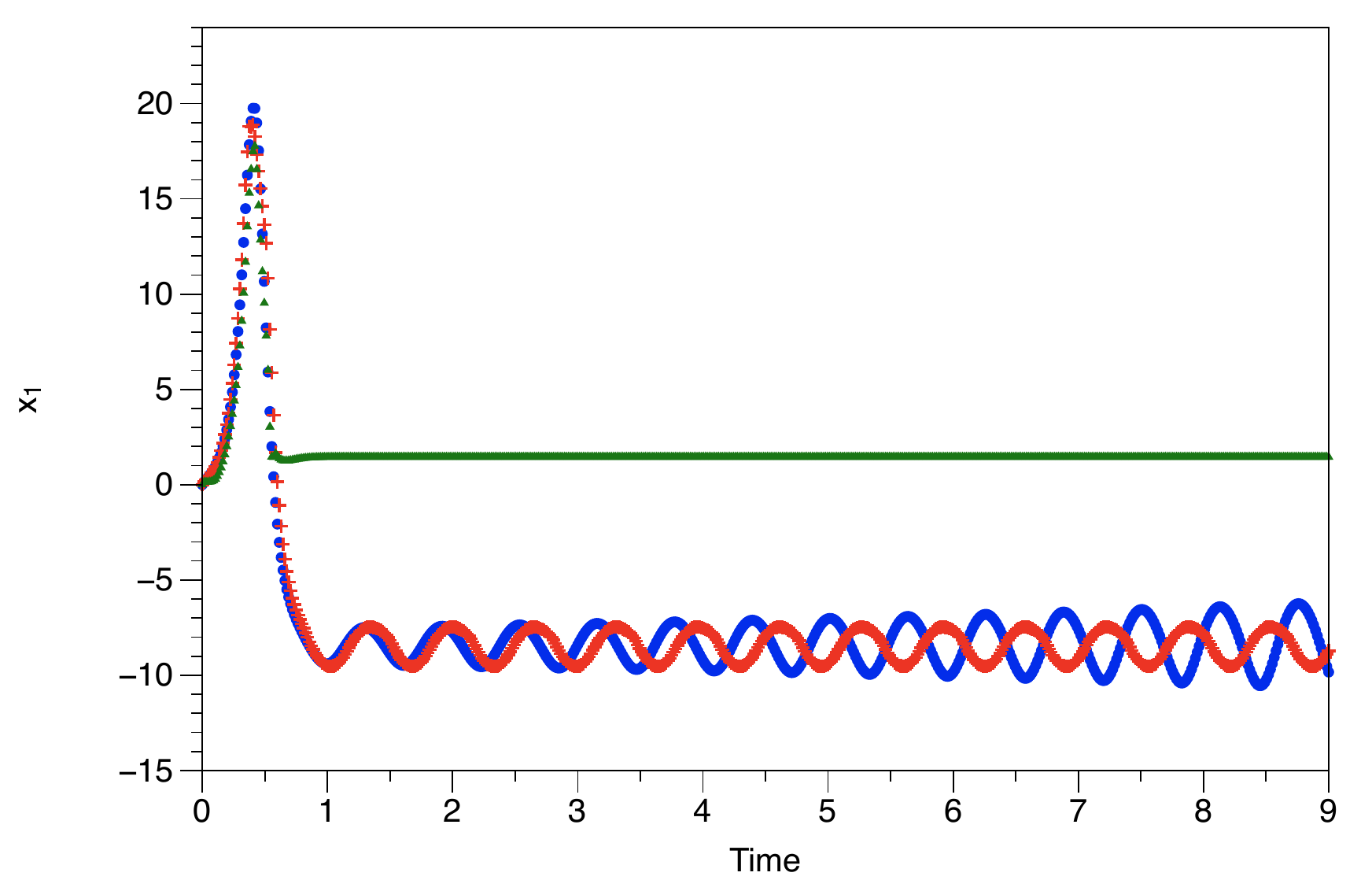}
   \label{plot_lorenz_supervised_noisevsnoiselessa}}
      \quad
   \subfigure[]{%
   \includegraphics[width = 5cm]{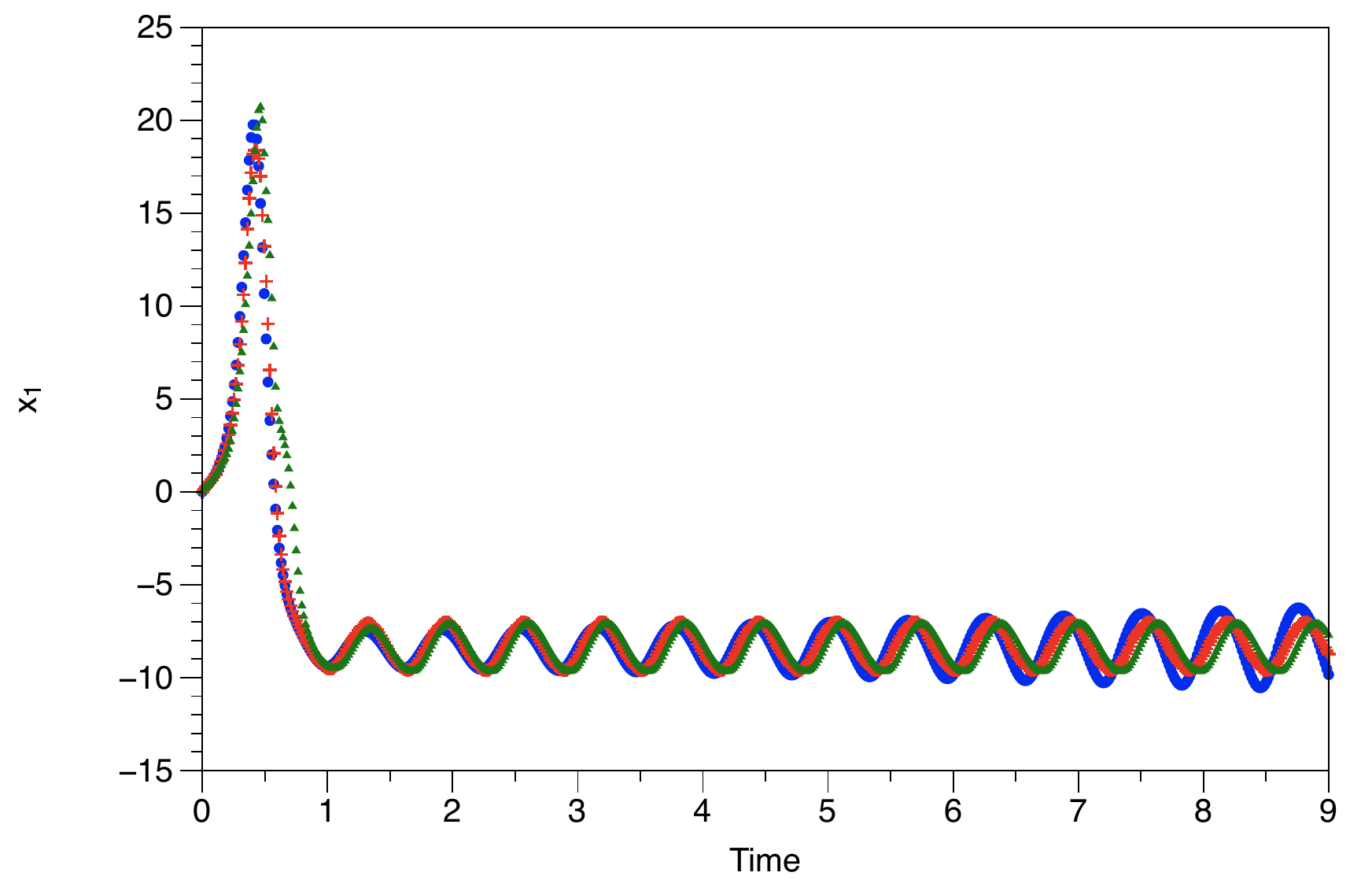}
   \label{plot_lorenz_supervised_noisevsnoiselessb}}
\caption{Supervised learning. Comparison of ground truth for $x_1(t)$ computed with the Euler scheme with timestep $\delta t=10^{-4}$ (blue dots), the neural network flow map prediction with timestep $\Delta t=1.5\times10^{-2}$ using {\it noisy} training data (red crosses) and the neural network flow map prediction with timestep $\Delta t=1.5\times10^{-2}$ using {\it noiseless} training data (green triangles). (a) {\it without enforced constraints} during training; (b) {\it with enforced constraints} during training (see text for details).  }
\label{plot_lorenz_supervised_noisevsnoiseless}
\end{figure}

Fig. \ref{plot_lorenz_supervised_noisevsnoiseless} compares the predictions of neural networks trained with noisy and noiseless training data. In addition, we perform such comparison both for the case {\it with enforced constraints} during training and {\it without enforced constraints}. Fig. \ref{plot_lorenz_supervised_noisevsnoiselessa} shows that when the constraints are {\it not} enforced during training, the use of noisy data can have a significant impact. This is along the lines of our argument that the data from a single training trajectory are not enough by themselves to train the neural network accurately for prediction purposes. Fig. \ref{plot_lorenz_supervised_noisevsnoiselessb} shows that when the constraints {\it are} enforced during training, the difference between the predictions based on noisy and noiseless training data is reduced. However, using noisy data results in better predictions for parts of the trajectory where there are rapid changes. Also the use of noisy data helps the prediction to stay ``in phase" with the ground truth for longer times. 

We have to stress that we conducted several numerical experiments and the performance of the neural network flow map trained with {\it noisy}  data was consistently more robust than when it was trained with {\it noiseless} data. A thorough comparison will appear in a future publication. 

\subsubsection{Error-correction for training with {\it noiseless} trajectory}

The results from Fig. \ref{plot_lorenz_supervised_noisevsnoiselessb} prompted us to examine in more detail the role of the error-correction term in the case of training with {\it noiseless} data. In particular, we would like to see how much of the predictive accuracy is due to enforcing the forward Euler scheme alone i.e. set $a_1=a_2=a_3=0$ in \eqref{lorenz_modified1}-\eqref{lorenz_modified3} versus allowing $a_1,a_2,a_3$ to be optimized during training. 

\begin{figure}[ht]
   \centering
   \subfigure[]{%
   \includegraphics[width = 5cm]{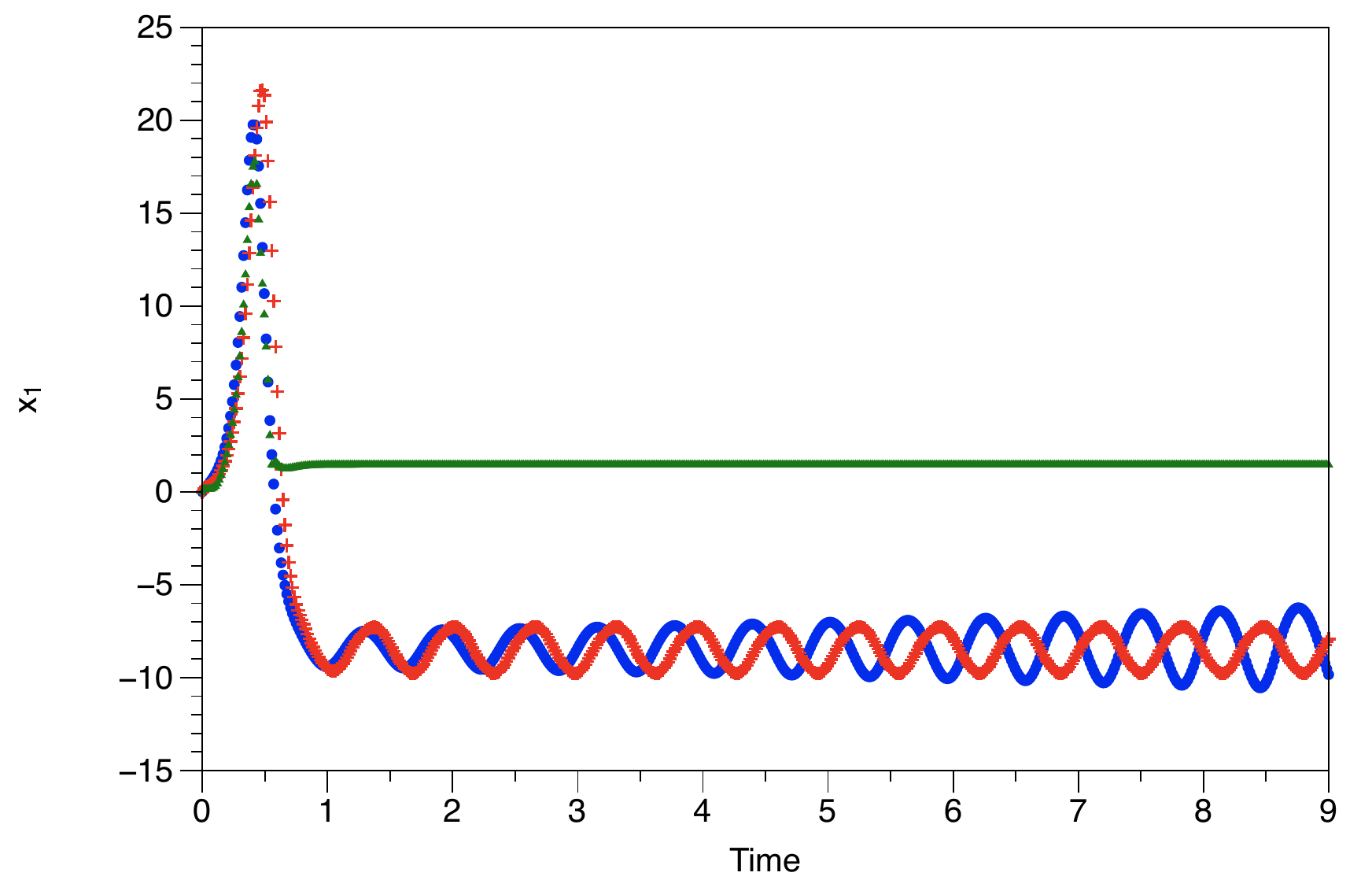}
   \label{plot_lorenz_supervised_noiselessa}}
      \quad
   \subfigure[]{%
   \includegraphics[width = 5cm]{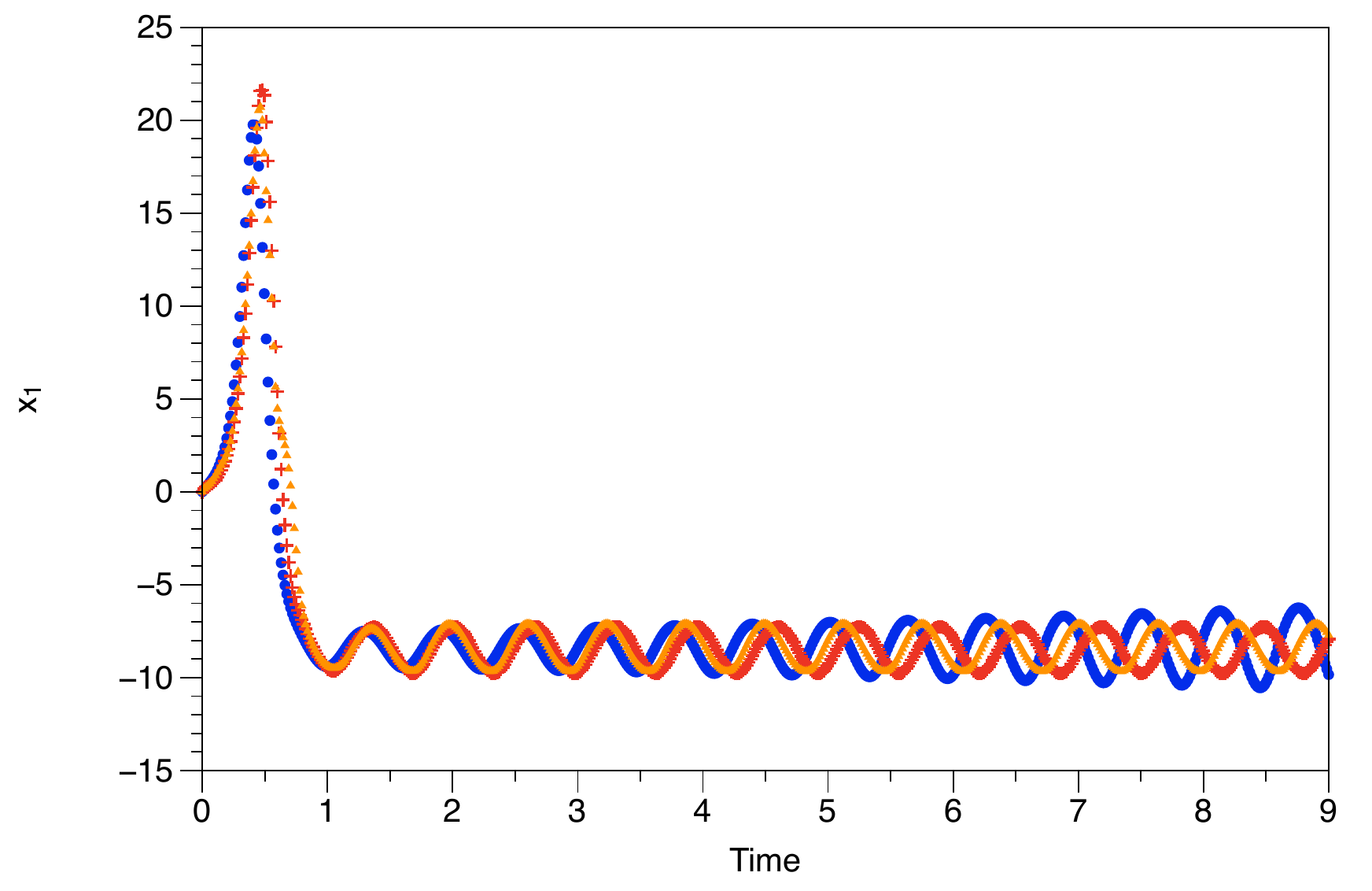}
   \label{plot_lorenz_supervised_noiselessb}}
\caption{Supervised learning. (a) Comparison of ground truth for $x_1(t)$ computed with the Euler scheme with timestep $\delta t=10^{-4}$ (blue dots), the neural network flow map prediction with timestep $\Delta t=1.5\times10^{-2}$ using {\it noiseless} training data and enforcing the Euler scheme {\it without} error-correction (red crosses) and the neural network flow map prediction with timestep $\Delta t=1.5\times10^{-2}$ using {\it noiseless} training data and {\it without} enforcing any constraints (green triangles); (b) Comparison of ground truth for $x_1(t)$ computed with the Euler scheme with timestep $\delta t=10^{-4}$ (blue dots), the neural network flow map prediction with timestep $\Delta t=1.5\times10^{-2}$ using {\it noiseless} training data and enforcing the Euler scheme {\it without} error-correction (red crosses) and the neural network flow map prediction with timestep $\Delta t=1.5\times10^{-2}$ using {\it noiseless} training data and enforcing the Euler scheme {\it with} error-correction (orange triangles)(see text for details).  }
\label{plot_lorenz_supervised_noiseless}
\end{figure}

Fig. \ref{plot_lorenz_supervised_noiselessa} compares to the ground truth the prediction from the trained neural network flow map when we do {\it not} enforce any constraints and the prediction from the trained neural network flow map when we enforce {\it only}  the forward Euler part of the constraints \eqref{lorenz_modified1}-\eqref{lorenz_modified3} ($a_1=a_2=a_3=0$). We see that indeed, even if we enforce only the forward Euler part of the constraint we obtain much more accurate results than not enforcing any constraint at all. 

Fig. \ref{plot_lorenz_supervised_noiselessb} examines how is the performance of the neural network affected further if we allow also the error-correcting term in \eqref{lorenz_modified1}-\eqref{lorenz_modified3} i.e. optimize $a_1,a_2,a_3$ during training. The inclusion of the error-correcting term allows the solution to remain for longer ``in phase" with the ground truth than if the error-correction term is absent. This is expected since we are examining the solution of the Lorenz system as it transitions from an initial condition far from the attractor to the attractor and then evolves on it. While on the attractor the solution remains oscillatory and bounded, so the main error of the neural network flow map prediction comes from going ``out of phase" with the ground truth. The error-correcting term keeps the predicted trajectory closer to the ground truth thus reducing the loss of phase. Recall that the error-correcting term is one of the simplest possible. From our prior experience with model reduction, we anticipate larger gains in accuracy if we use more sophisticated error-correcting terms.    

We want to stress again that training with {\it noiseless} data is significantly less robust than training with {\it noisy} data. However, we have chosen to present results of training with {\it noiseless} data that exhibit good prediction accuracy to raise various issues that should be more thoroughly investigated.

\subsection{Unsupervised learning}\label{numerical_unsupervised}

\begin{figure}[ht]
   \centering
   \subfigure[]{%
   \includegraphics[width = 5cm]{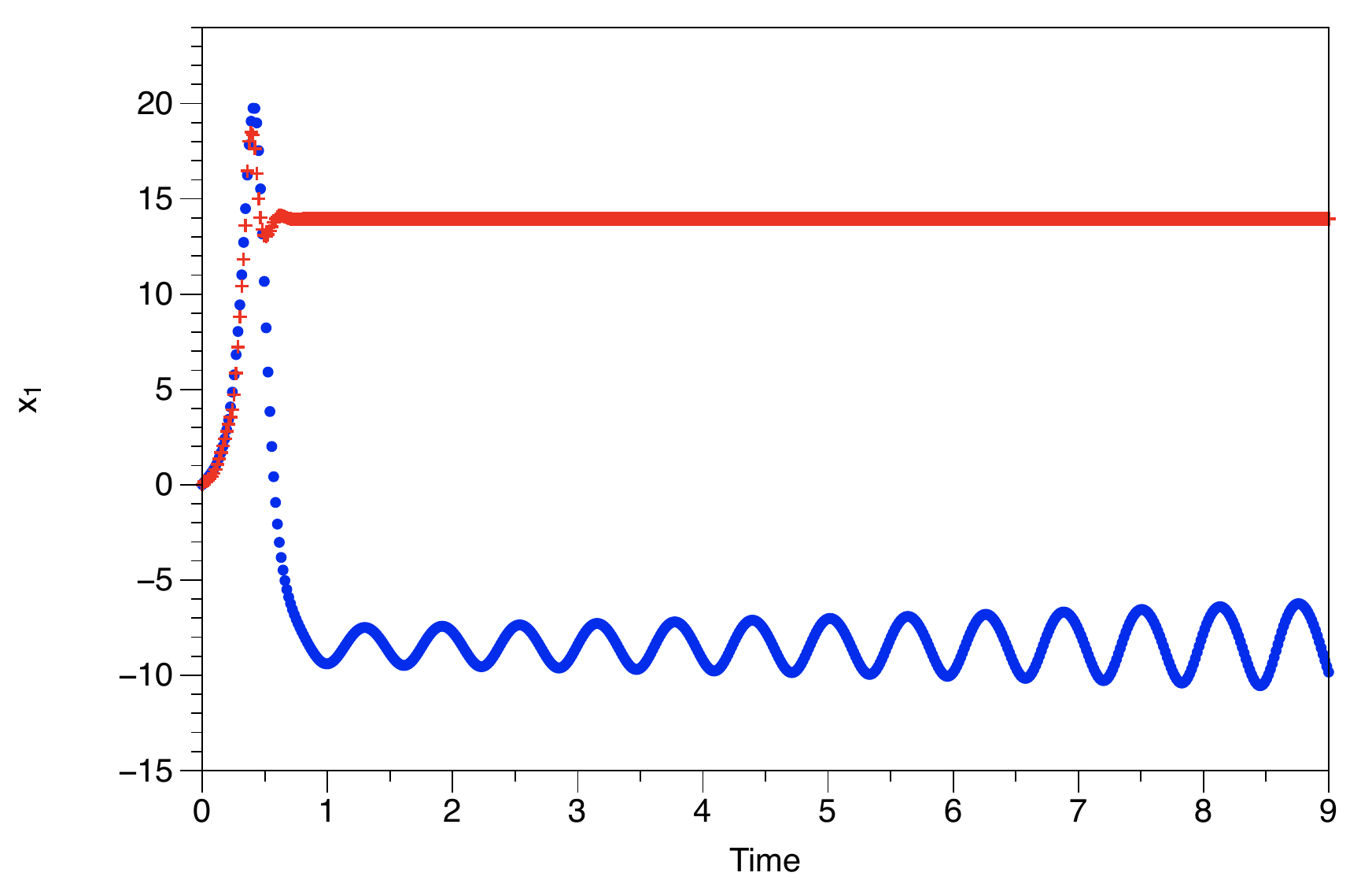}
   \label{plot_lorenz_unsuperviseda}}
      \quad
   \subfigure[]{%
   \includegraphics[width = 5cm]{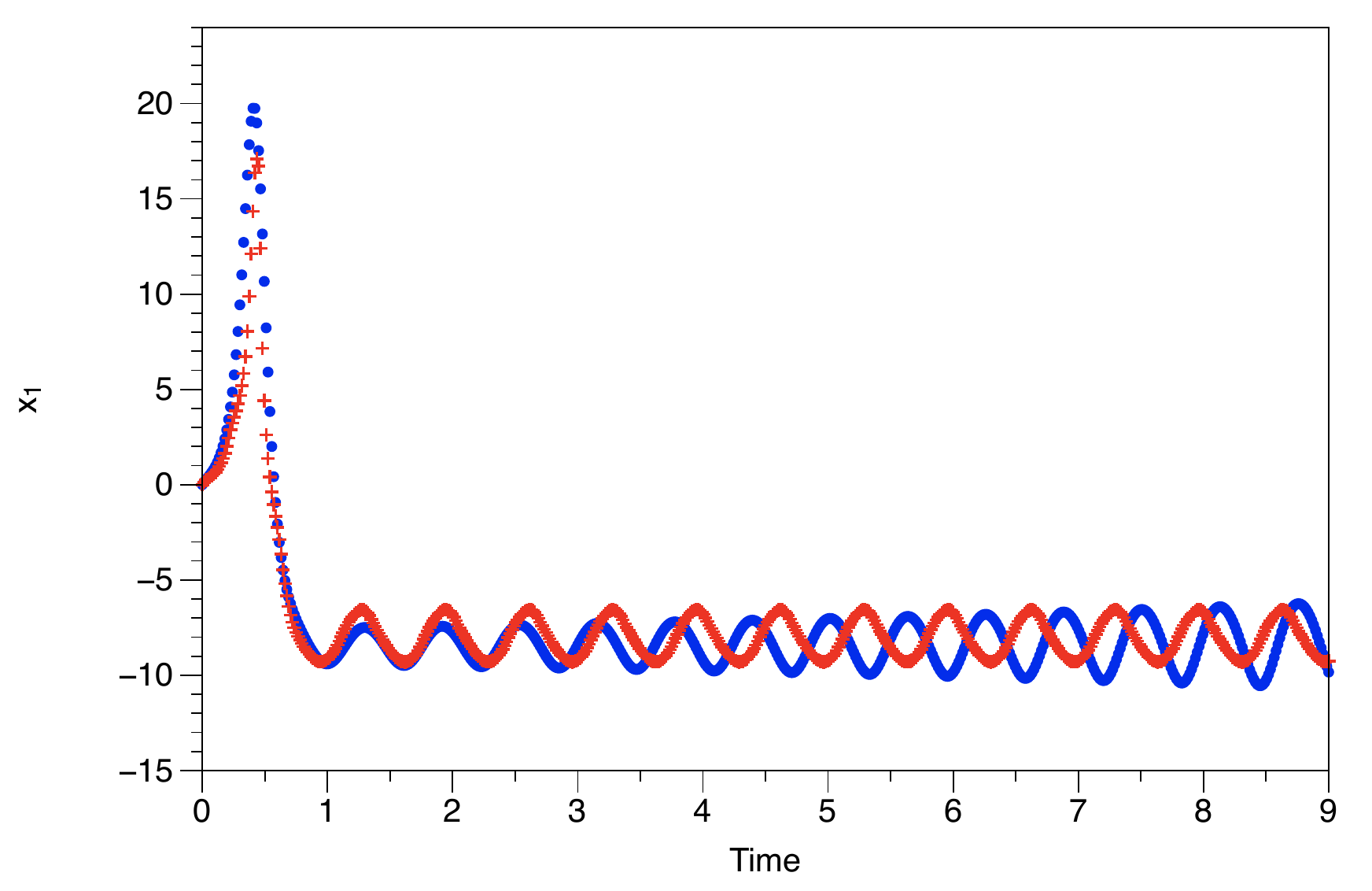}
   \label{plot_lorenz_unsupervisedb}}
\caption{Unsupervised learning (GAN). Comparison of ground truth for $x_1(t)$ computed with the Euler scheme with timestep $\delta t=10^{-4}$ (blue dots) and the neural network flow map (GAN generator) prediction with timestep $\Delta t=1.5\times10^{-2}$ (red crosses). (a) {\it noisy} data {\it without enforced constraints} during training; (b) {\it noisy} data {\it with enforced constraints} during training (see text for details).  }
\label{plot_lorenz_unsupervised}
\end{figure}

We continue with the case of {\it unsupervised} learning and in particular the case of a GAN. We have used for the GAN generator a deep neural network with 9 hidden layers of width 20 and for the discriminator a neural network with 2 hidden layers of width 20. The numbers of hidden layers both for the generator and the discriminator were chosen as the smallest that allowed the GAN training to reach its game-theoretic optimum without at the same time requiring large scale computations. Fig. \ref{plot_lorenz_unsupervised} compares the evolution of the prediction of the neural network flow map starting at $t=0$ and computed with a timestep $\Delta t=1.5\times10^{-2}$ to the ground truth (training trajectory) computed with the forward Euler scheme with timestep $\delta t=10^{-4}.$  

Fig. \ref{plot_lorenz_unsuperviseda} shows results for the {\it implicit} enforcing of constraints. We see that this is not enough to produce a neural network flow map with long-term predictive accuracy. Fig. \ref{plot_lorenz_unsupervisedb} shows the significant improvement in the predictive accuracy when we enforce the constraints {\it explicitly}. The results for this specific example are not as good as in the case of supervised learning presented earlier. We note that training a GAN with or without constraints is a delicate numerical task as explained in more detail in \cite{stinis2018}. One needs to find the right balance between the expressive strengths of the generator and the discriminator (game-theoretic optimum) to avoid instabilities but also train the neural network flow map i.e. the GAN generator, so that it has predictive accuracy. 

We also note that training with {\it noiseless} data is even more brittle. For the very few experiments where we avoided instability the predicted solution from the trained GAN generator was not accurate at all.

\subsection{Reinforcement learning}\label{numerical_reinforcement}

The last case we examine is that of {\it reinforcement} learning. In particular, we want to see how an actor-critic method performs in the difficult case when the discount factor $\gamma=1.$ We repeat that $\gamma=1$ corresponds to the case of a deterministic environment which means that the same actions always produce the same rewards. This is the situation in our numerical experiments where we are given a training trajectory that does not change. We have conducted more experiments for other values of $\gamma$ but a detailed presentation of those results will await a future publication. 

For the representation of the action-value function we used a deep neural network with 15 hidden layers of width 20. For the representation of the deterministic action policy i.e. the neural network flow map in our parlance, we used a deep neural network with 10 hidden layers of width 20. The task of learning an accurate representation of the action-value function is more difficult than that of finding the action policy. This justifies the need for a stronger network to represent the action-value function.

As we have mentioned Section \ref{reinforcement}, researchers have developed various modifications and tricks to stabilize the training of AC methods \cite{pfau2016}. The one that enabled us to stabilize results in the first place is that of {\it target networks} \cite{mnih2015,lillicrap2015}. However, the predictive accuracy of the trained neural network flow map i.e. the action policy, was extremely poor unless we {\it also} used our homotopy approach for the action-value function. This was true for both cases of enforcing or not constraints explicitly during training. With this in mind we present results with and without the homotopy approach for the action-value function to highlight the accuracy improvement afforded by the use of homotopy.

Before we present the results we provide some details about the target networks, the reward function and the specifics of the homotopy schedule. The target network concept uses different networks to represent the action-value function and the action policy that appear in the expression for the target \eqref{bellman_opt_target_off}. In particular, if $\theta_Q$ and $\theta_{\mu}$ are the parameter vectors for the action-value function and action policy respectively, then we use neural networks with parameter vectors $\theta_{Q'}$ and $\theta_{\mu'}$ (the {\it target networks}) to evaluate the target expression \eqref{bellman_opt_target_off}. The vectors $\theta_{Q'}$ and $\theta_{\mu'}$ can be initialized with the same values as $\theta_Q$ and $\theta_{\mu}$ but they evolve in a different way. In fact, after every iteration update for $\theta_Q$ and $\theta_{\mu}$ we apply the update rule
\begin{align}
\theta_{Q'} & \leftarrow  \tau \theta_{Q} + (1-\tau) \theta_{Q'} \\
\theta_{\mu'} & \leftarrow  \tau \theta_{\mu} + (1-\tau) \theta_{\mu'}
\end{align}   
where we have taken $\tau=0.001$ \cite{lillicrap2015}. 

The reward function (with constraints) for an input point $z$ from the noise cloud
\begin{gather}
r(z,x)=-\bigg \{  \sum_{j=1}^{3} \bigg[ (\mu_j(z)-x_j^{data})^2 \bigg] \\
+ (\mu_1(z)-z_{1} - \Delta t [\sigma (z_{2}-z_{1})] + \Delta t a_1  z_{1})^2  \\
+ (\mu_2(z)-z_{2} - \Delta t [\rho z_{1} - z_{2}- z_{1} z_{3}] +  \Delta t a_2  z_{2})^2 \\
+ (\mu_3(z)-z_{3} - \Delta t [ z_{1} z_{2} - \beta z_{3}] + \Delta t a_3  z_{3} )^2 \bigg \}
\end{gather}
where $x^{data}$ is the {\it noiseless} point from the training trajectory. As we have explained in Section \ref{reinforcement} (see the comment after \eqref{bellman_opt_target_off}), in the AC method context the distribution of the noise cloud of the input data points at every timestep corresponds to the state visitation distribution $\rho^{\beta}$ appearing in \eqref{bellman_opt_off}.

\begin{figure}[ht]
   \centering
   \subfigure[]{%
   \includegraphics[width = 5cm]{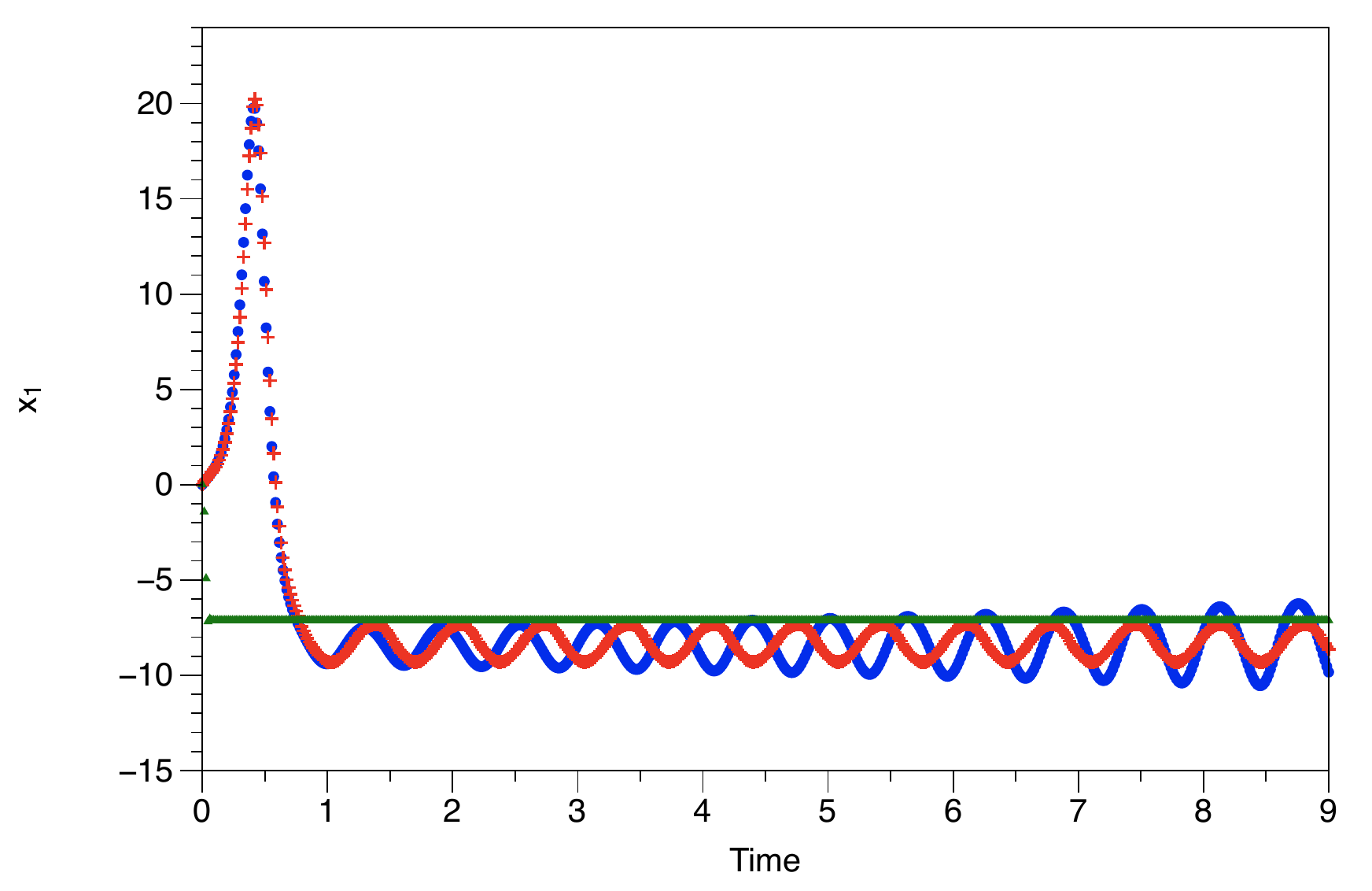}
   \label{plot_lorenz_reinforcementa}}
      \quad
   \subfigure[]{%
   \includegraphics[width = 5cm]{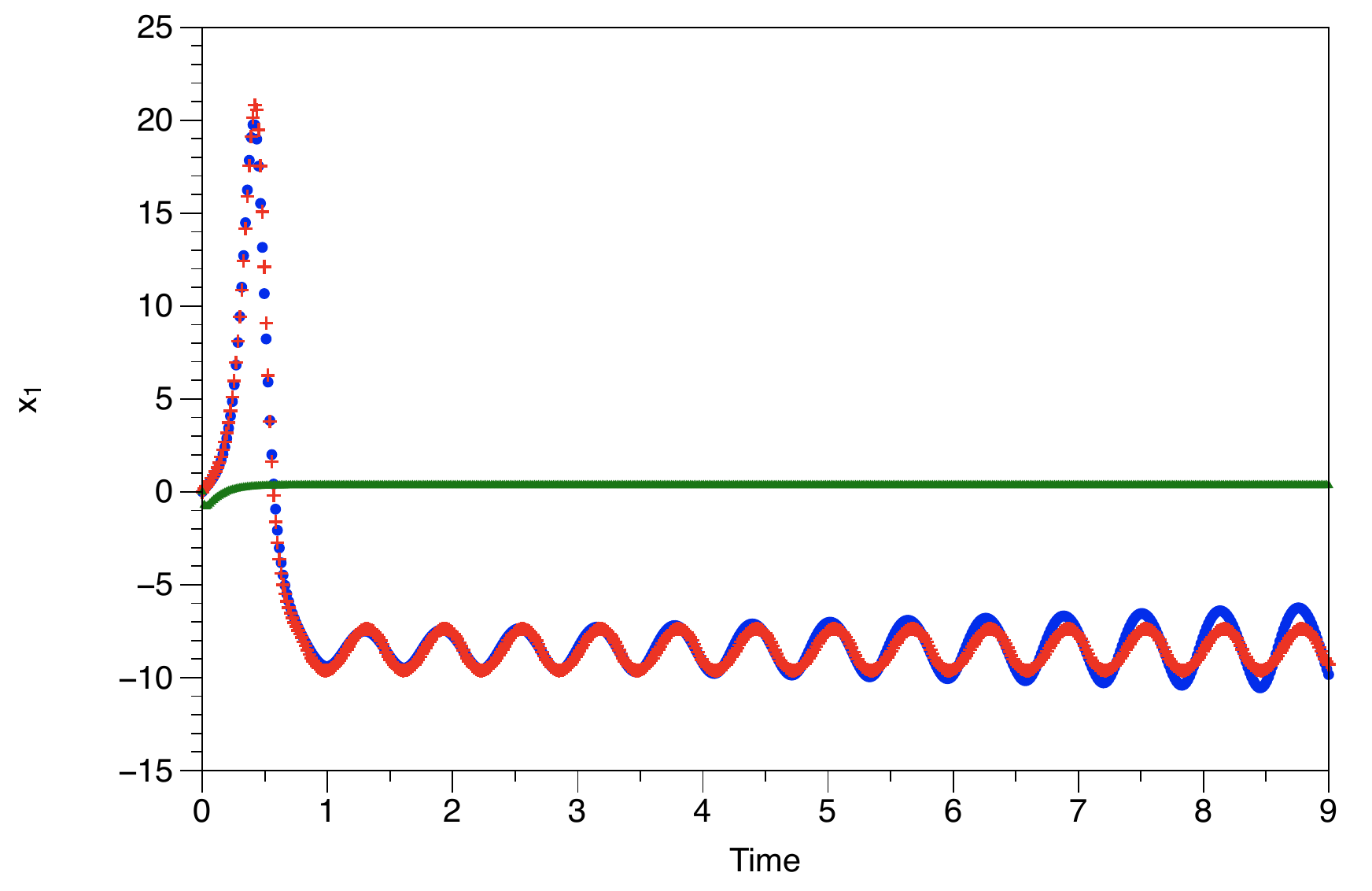}
   \label{plot_lorenz_reinforcementb}}
\caption{Reinforcement learning (Actor-critic). Comparison of ground truth for $x_1(t)$ computed with the Euler scheme with timestep $\delta t=10^{-4} $ (blue dots), the neural network flow map prediction with timestep $\Delta t=1.5\times10^{-2}$ {\it with} homotopy for the action-value function during training (red crosses) and the neural network flow map prediction with timestep $\Delta t=1.5\times10^{-2}$ {\it without} homotopy for the action-value function during training (green triangles). (a) {\it noisy} data {\it without enforced constraints} during training; (b) {\it noisy} data {\it with enforced constraints} during training (see text for details).  }
\label{plot_lorenz_reinforcement}
\end{figure}

The homotopy schedule we used is a rudimentary one that we did not attempt to optimize. Obviously, this is a topic of further investigation that will appear elsewhere. We initialized the homotopy parameter $\delta$ at 0, and increased its value (until it reached 1) every 2000 iterations of the optimization. 

Fig. \ref{plot_lorenz_reinforcement} presents results of the prediction performance of the neural network flow map when it was trained with and without the use of homotopy for the action value function. In Fig. \ref{plot_lorenz_reinforcementa} we have results for the {\it implicit} enforcing of constraints while in Fig. \ref{plot_lorenz_reinforcementb} for the {\it explicit} enforcing of constraints. We make two observations. First, both for implicit and explicit enforcing of the constraints, the use of homotopy leads to accurate results for long times. Especially for the case of explicit enforcing which gave us some of the best results from all the numerical experiments we conducted for the different modes of learning. Second, if we do not use homotopy, the predictions are extremely poor both for implicit and explicit forcing. Indeed, the green curve in Fig. \ref{plot_lorenz_reinforcementa} representing the prediction of $x_1(t)$ for the case of implicit constraint enforcing {\it without} homotopy is as inaccurate as it looks. It starts at 0 and within a few steps drops to a negative value and does not change much after that. The predictions for $x_2(t)$ and $x_3(t)$ are equally inaccurate.


\section{Discussion and future work}\label{discussion}

We have presented a collection of results about the enforcing of known constraints for a dynamical system during the training of a neural network to represent the flow map of the system. We have provided ways that the constraints can be enforced in all three major modes of learning, namely supervised, unsupervised and reinforcement learning. In line with the law of scientific computing that one should build in an algorithm as much prior information is known as possible, we observe a striking improvement in performance when known constraints are enforced during training. We have also shown the benefit of training with noisy data and how these correspond to the incorporation of a restoring force in the dynamics of the system. This restoring force is analogous to memory terms appearing in model reduction formalisms. In our framework, the reduction is in a {\it temporal} sense i.e. it allows us to construct a flow map that remains accurate though it is defined for {\it large} timesteps. 

The model reduction connection opens an interesting avenue of research that makes contact with complex systems appearing in real-world problems. The use of larger timesteps for the neural network flow map than the ground truth without sacrificing too much accuracy is important. We can imagine an online setting where observations come at {\it sparsely} placed time instants and are used to update the parameters of the neural network flow map. The use of sparse observations could be dictated by {\it necessity} e.g. if it is hard to obtain frequent measurements or {\it efficiency} e.g. the local processing of data in field-deployed sensors can be costly. Thus, if the trained flow map is capable of accurate estimates using {\it larger} timesteps then its successful updated training using only sparse observations becomes more probable.       

The constructions presented in the current work depend on a large number of details that can potentially affect their performance. A thorough study of the relative merits of enforcing constraints for the different modes of learning needs to be undertaken and will be presented in a future publication. We do believe though that the framework provides a promising research direction at the nexus of scientific computing and machine learning. 


\section{Acknowledgements}
The author would like to thank Court Corley, Tobias Hagge, Nathan Hodas, George Karniadakis, Kevin Lin, Paris Perdikaris, Maziar Raissi, Alexandre Tartakovsky, Ramakrishna Tipireddy, Xiu Yang and Enoch Yeung for helpful discussions and comments. The work presented here was partially supported by the PNNL-funded ``Deep Learning for Scientific Discovery Agile Investment" and the DOE-ASCR-funded "Collaboratory on Mathematics and Physics-Informed Learning Machines for Multiscale and Multiphysics Problems (PhILMs)". Pacific Northwest National Laboratory is operated by Battelle Memorial Institute for DOE under Contract DE-AC05-76RL01830.

\bibliographystyle{plain}
\bibliography{theory}

\end{document}